\documentclass[11pt,a4paper]{article}
\usepackage[hyperref]{acl2017}
\usepackage{times}
\usepackage{latexsym}

\usepackage{url}

\usepackage[pdftex]{graphicx} 
\usepackage{amsmath} 
\usepackage{epstopdf}
\usepackage{CJK} 
\usepackage{multirow} 
\usepackage{array}

\usepackage{algorithm}
\usepackage{algpseudocode}

\aclfinalcopy 

\title{Neural System Combination for Machine Translation}

\author{Long Zhou$^{\dagger}$, Wenpeng Hu$^{\dagger}$, Jiajun Zhang$^{\dagger}$\thanks{\ \ Corresponding author.},
  Chengqing Zong$^{\dagger\ddagger}$ \\
  $^\dagger$University of Chinese Academy of Sciences, Beijing, China \\
  National Laboratory of Pattern Recognition, CASIA, Beijing, China \\
  $^\ddagger$CAS Center for Excellence in Brain Science and Intelligence Technology, Shanghai, China \\
  {\tt \{long.zhou,wenpeng.hu,jjzhang,cqzong\}@nlpr.ia.ac.cn} \\}

\date{}

\begin{document}

\begin{CJK*}{UTF8}{gbsn}

\maketitle
\begin{abstract}
  Neural machine translation (NMT) becomes a new approach to machine translation and generates much more fluent
  results compared to statistical machine translation (SMT).
  However, SMT is usually better than NMT in translation adequacy.
  It is therefore a promising direction to combine the advantages of both NMT and SMT.
  In this paper, we propose a neural system combination framework leveraging multi-source NMT,
  which takes as input the outputs of NMT and SMT systems and produces the final translation.
  Extensive experiments on the Chinese-to-English translation task show that our model archives significant
  improvement by 5.3 BLEU points over the best single system output and 3.4 BLEU points over the state-of-the-art
  traditional system combination methods.

\end{abstract}

\section{Introduction}

Neural machine translation has significantly improved the quality of machine translation in recent several years
~\cite{Kalchbrenner:2013,Sutskever:2014,Bahdanau:2015,Junczys-Dowmunt:2016A}.
Although most sentences are more fluent than translations by statistical machine translation (SMT)~\cite{Koehn:2003,Chiang:2005},
NMT has a problem to address translation adequacy especially for the rare and unknown words. Additionally,
it suffers from over-translation and under-translation to some extent~\cite{Tu:2016}.
Compared to NMT, SMT, such as phrase-based machine translation (PBMT, \cite{Koehn:2003}) and hierarchical phrase-based machine translation
(HPMT, \cite{Chiang:2005}), does not need to limit the vocabulary and can guarantee translation coverage of source sentences.
It is obvious that NMT and SMT have different strength and weakness. In order to take full advantages of both NMT and SMT, system combination can be a good choice.

Traditionally, system combination has been explored respectively in sentence-level, phrase-level, and word-level ~\cite{Kumar:2004,Feng:2009,Chen:2009}. Among them, word-level combination approaches that adopt confusion network for
decoding have been quite successful~\cite{Rosti:2007,Ayan:2008,Freitag:2014}. However, these approaches are mainly designed for SMT without considering the features of NMT results.
NMT opts to produce diverse words and free word order, which are quite different from SMT.
And this will make it hard to construct a consistent confusion network.
Furthermore, traditional system combination approaches cannot guarantee the fluency of the final translation results.

In this paper, we propose a neural system combination framework, which is adapted from the multi-source NMT model {\cite{Zoph:2016}}.
Different encoders are employed to model the semantics of the source language input and each best translation produced by different NMT and SMT systems. The encoders produce multiple context vector representations, from which the decoder generates the final output word by word.
Since the same training data is used for NMT, SMT and neural system combination, we further design a smart strategy to simulate the real training data for neural system combination.

\begin{figure*}
    \centering
    \includegraphics[height=7cm]{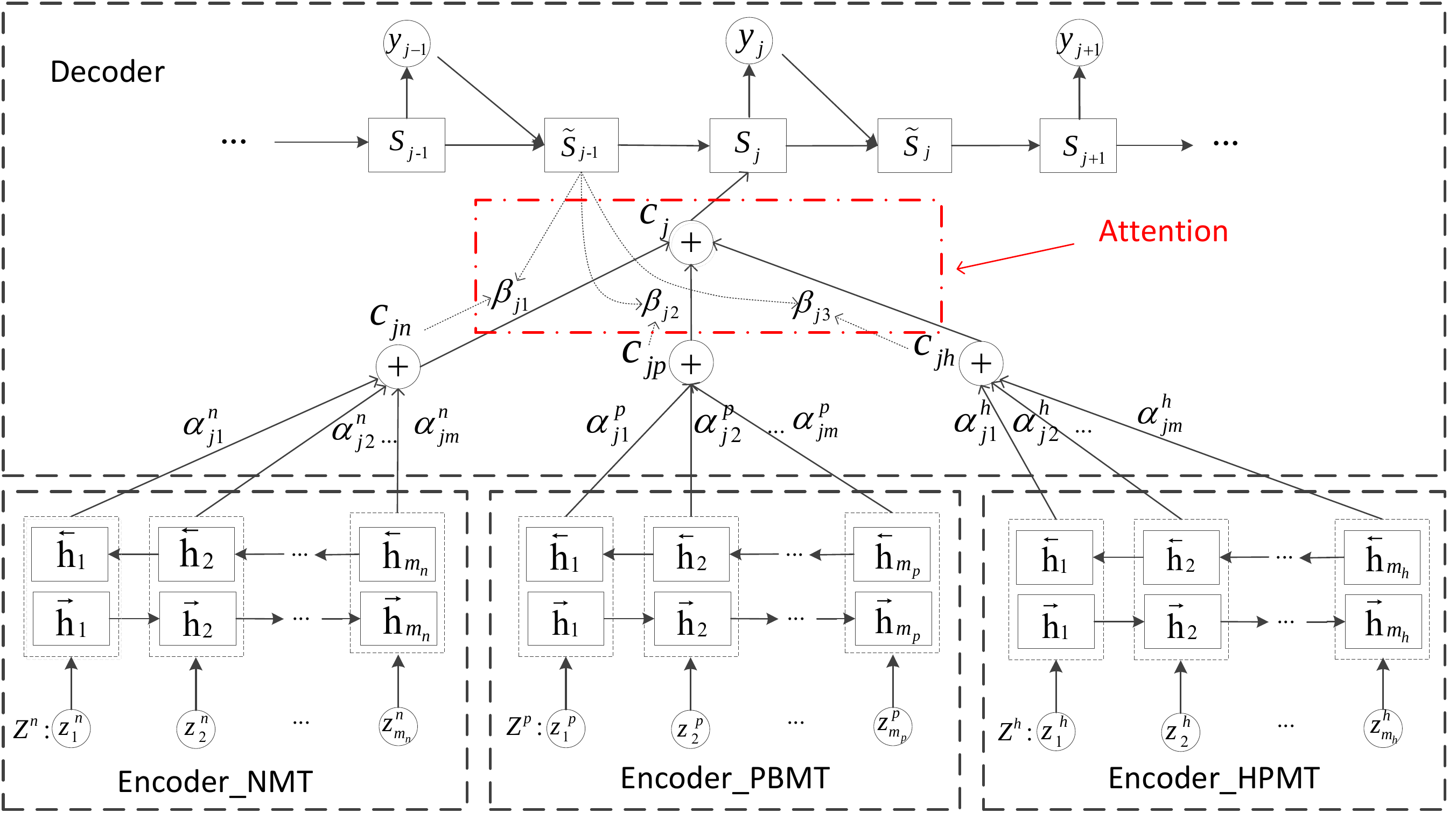}
    \caption{The architecture of Neural System Combination Model.}\label{fig:2}
\end{figure*}

Specifically, we make the following contributions in this paper:
\begin{itemize}
\item We propose a neural system combination method, which is adapted from multi-source NMT model and can accommodate both source inputs and different system translations. It combines the fluency of NMT and adequacy (especially the ability to address rare words) of SMT.
\item We design a good strategy to construct appropriate training data for neural system combination.
\item The extensive experiments on Chinese-English translation show that our model archives significant improvement by 3.4 BLEU points over the state-of-the-art system combination methods and 5.3 BLEU points over the best individual system output.
\end{itemize}

\section{Neural Machine Translation}

The encoder-decoder NMT with an attention mechanism~\cite{Bahdanau:2015} has been proposed to softly align each decoder state with the encoder states,
and computes the conditional probability of the translation given the source sentence.

The encoder is a bidirectional neural network with gated recurrent units (GRU)~\cite{Cho:2014} which reads an input sequence $X=(x_1,x_2,...,x_m)$
and encodes it into a sequence of hidden states $H=h_1,h_2,...,h_m$.

The decoder is a recurrent neural network that predicts a target sequence $Y=(y_1,y_2,...,y_n)$. Each word $y_j$ is predicted based on a recurrent hidden
state $s_j$, the previously predicted word $y_{j-1}$, and a context vector $c_j$. $c_j$ is obtained from the weighted sum of the annotations $h_i$.
We use the latest implementation of attention-based NMT\footnote{https://github.com/nyu-dl/dl4mt-tutorial}.

\section{Neural System Combination for Machine Translation}

Macherey and Och \shortcite{Macherey:2007} gave empirical evidence that these systems to be combined need to be
almost uncorrelated in order to be beneficial for system combination. Since NMT and SMT are two kinds of
translation models with large differences, we attempt to build a neural system combination model, which can take advantage of the different systems.

{\bf Model:} Figure 1 illustrates the neural system combination framework, which can take as input the source sentence and the results of MT systems.
Here, we use MT results as inputs to detail the model.

Formally, given the result sequences $Z$($Z^n$, $Z^p$, and $Z^h$) of three MT systems for the same source sentence and previously generated target sequence
$Y_{<j}=(y_1,y_2,...,y_{j-1})$,
the probability of the next target word $y_j$ is
\begin{equation}
    p(y_j|Y_{<j},Z) = softmax(f(c_j,y_{j-1},s_j))
\end{equation}
Here $f(.)$ is a non-linear function, $y_{j-1}$ represents the word embedding of the previous prediction word, and $s_j$ is the state of decoder
at time step j, calculated by

\begin{table*}
\centering
\begin{tabular}{lccccc}
  \hline
  System   &    MT03   &   MT04   &  MT05   &   MT06   & Ave \\
  \hline
  \hline
  PBMT     &   37.47    &   41.20   &  36.41   &   36.03   &   37.78   \\
  HPMT     &   {\bf 38.05}     &   {\bf 41.47}   & {\bf 36.86}   &   36.04   &  {\bf 38.10}   \\
  NMT      &   37.91     &   38.95   &  36.02   &   {\bf 36.65}   &   37.38   \\
  \hline
  Jane~\cite{Freitag:2014}    &   39.83     &   42.75   &  38.63   &   39.10   &   40.08   \\
  Multi    &   40.64     &   44.81   &  38.80   &   38.26   &   40.63   \\
  Multi+Source &   42.16   &   45.51   &  40.28   &   39.03   &   41.75    \\

  Multi+Ensemble    &   41.67     &   45.95   &  40.37   &   39.02   &   41.75   \\
  Multi+Source+Ensemble &   {\bf 43.55}   &   {\bf 47.09}   &  {\bf 42.02}   &   {\bf 41.10}   &   {\bf 43.44}    \\
  \hline
\end{tabular}
\caption{Translation results (BLEU score) for different machine translation and system combination methods.
  Jane is a open source machine translation system combination toolkit that uses confusion network decoding.
  {\bf Best} and {\bf important} results per category are highlighted.
  }
\end{table*}

\begin{equation}
    s_j = GRU(\tilde{s}_{j-1},c_j)
\end{equation}
\begin{equation}
    \tilde{s}_{j-1} = GRU(s_{j-1},y_{j-1})
\end{equation}
where $s_{j-1}$ is previous hidden state, $\tilde{s}_{j-1}$ is an intermediate state. And $c_j$ is the context vector of system combination obtained by attention mechanism,
which is computed as weighted sum of the context vectors of three MT systems,  just as illustrated in the middle part of Figure 1.
\begin{equation}
    c_j = \sum_{k=1}^{K}\beta_{jk}c_{jk}
\end{equation}
where K is the number of MT systems, and $\beta_{jk}$ is a normalized item calculated as follows:
\begin{equation}
    \beta_{jk} = \frac{exp(\tilde{s}_{j-1} \cdot c_{jk})} {\sum_{k^{'}} exp(\tilde{s}_{j-1} \cdot c_{jk^{'}})}
\end{equation}
Here, we calculate kth MT system context $c_{jk}$ as a weighted sum of the source annotations:
\begin{equation}
    c_{jk} = \sum_{i=1}^m\alpha^k_{ji}h_i
\end{equation}
where $h_i = [\overrightarrow{h}_i;\overleftarrow{h}_i]$ is the annotation of $z_i$ from a bi-directional GRU, and its weight $\alpha^k_{ji}$ is computed by
\begin{equation}
    \alpha^{k}_{ji} = \frac{exp(e_{ji})}{\sum_{l=1}^mexp(e_{jl})}
\end{equation}
where $e_{ji} = v_a^Ttanh(W_a\tilde{s}_{j-1}+U_ah_i)$ scores how well $\tilde{s}_{j-1}$ and $h_i$ match.


{\bf Training Data Simulation:} The neural system combination framework should be trained on the outputs of multiple translation systems and the gold target translations.
In order to keep consistency in training and testing, we design a strategy to simulate the real scenario. We randomly divide the training corpus into two parts, then reciprocally train the
MT system on one half and translate the source sentences of the other half into target translations. The MT translations and the gold target reference can be available.

\section{Experiments}

We perform our experiments on the Chinese-English translation task.
The MT systems participating in system combination are PBMT, HPMT and NMT.
The evaluation metric is case-insensitive BLEU ~\cite{Papineni:2002}.

\subsection{Data preparation}

Our training data consists of 2.08M sentence pairs extracted from LDC corpus.
We use NIST 2003 Chinese-English dataset as the validation set,
NIST 2004-2006 datasets as test sets.
We list all the translation methods as follows:
\begin{itemize}
  \item {\bf PBMT}: It is the start-of-the-art phrase-based SMT system. We use its default setting and train a 4-gram language model on the target portion of the bilingual training data.
  \item {\bf HPMT}: It is a hierarchical phrase-based SMT system, which uses its default configuration as PBMT in Moses.
  \item {\bf NMT}: It is an attention-based NMT system, with the same setting given in section 2.
\end{itemize}

\subsection{Training Details}

The hyper-parameters used in our neural combination system are described as follows.
We limit both Chinese and English vocabulary to 30k in our experiments.
The number of hidden units is 1000 and the word embedding dimension is 500 for all source and target word.
The network parameters are updated with Adadelta algorithm.
We adopt beam search with beam size b=10 at test time.

As to confusion-network-based system Jane, we use its default configuration and train a 4-gram language model on target data
and 10M Xinhua portion of Gigaword corpus.

\subsection{Main Results}

We compare our neural combination system with the best individual engines, and the state-of-the-art traditional combination
system Jane~\cite{Freitag:2014}. Table 1 shows the BLEU of different models on development data and test data.
The BLEU score of the multi-source neural combination model is 2.53 higher than the best single model HPMT.
The source language input gives a further improvement of +1.12 BLEU points.

As shown in Table 1, Jane outperforms the best single
MT system by 1.92 BLEU points. However, our neural combination system with source language gets an improvement of 1.67 BLEU points over Jane.
Furthermore, when augmenting our neural combination system with ensemble decoding {\footnote{We use four neural combination models in ensemble model.}},
it leads to another significant boost of +1.69 BLEU points.

%

\subsection{Word Order of Translation}

\begin{figure}
    \centering
    \includegraphics[width=7.3cm]{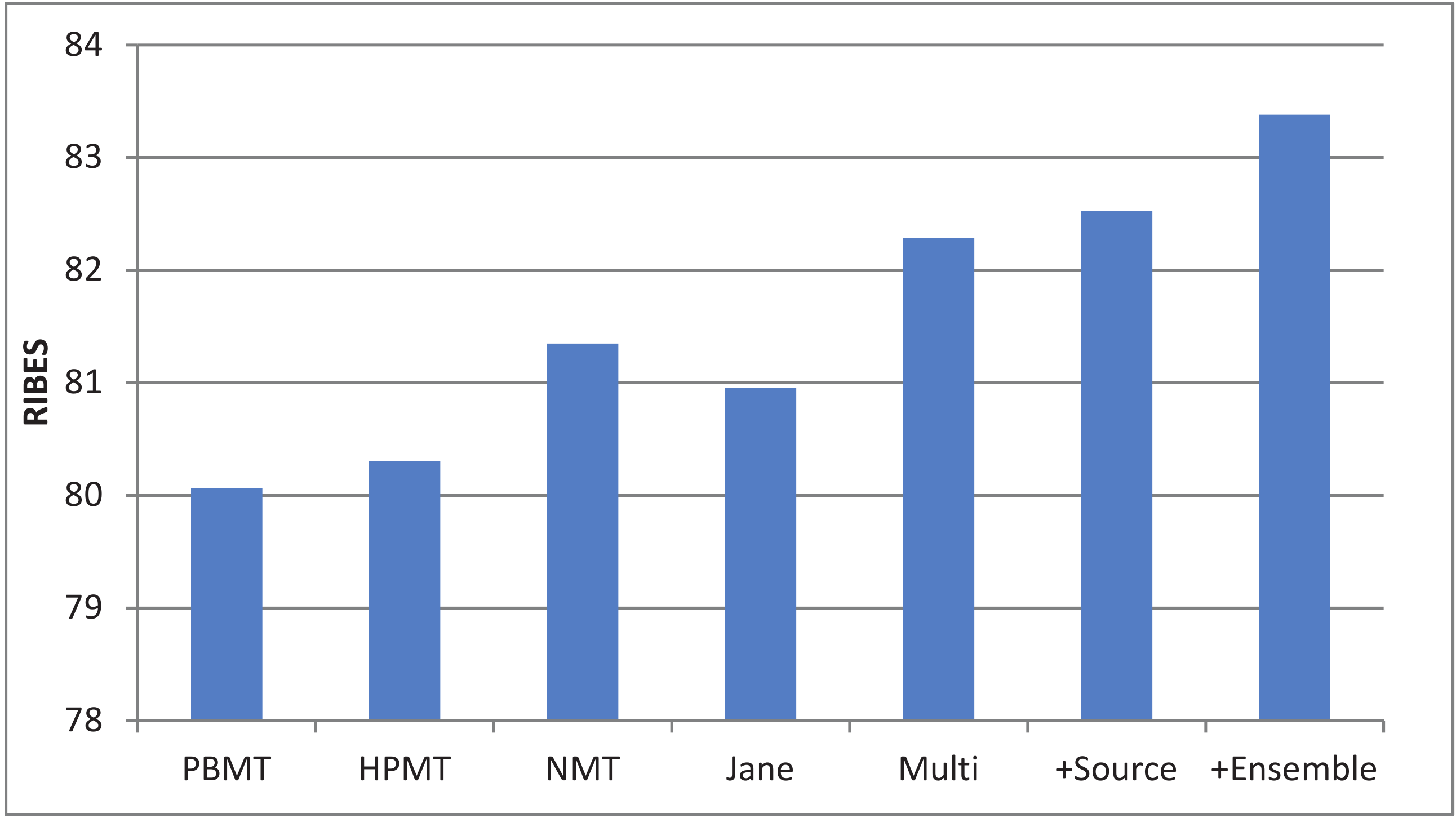}
    \caption{Translation results (RIBES score) for different machine translation and system combination methods.}\label{fig:4}
\end{figure}


We evaluate word order by the automatic evaluation metrics RIBES~\cite{Isozaki:2010}, whose score is a metric
based on rank correlation coefficients with word precision. RIBES is known to have stronger correlation with human judgements
than BLEU for English as discussed in Isozaki et al.~\shortcite{Isozaki:2010}.

Figure 2 illustrates experimental results of RIBES scores, which demonstrates that our
neural combination model outperforms the best result of single MT system and Jane.
Additionally, although BLEU point of Jane is higher than single NMT system, the word order of Jane is worse in terms of RIBES.

\subsection{Rare and Unknown Words Translation}

It is difficult for NMT systems to handle rare words, because low-frequency words in training data cannot capture latent translation mappings in neural network model.
However, we do not need to limit the vocabulary in SMT, which are often able to translate rare words in training data.
As shown in Table 2, the number of unknown words of our proposed model is 137 fewer than original NMT model.

Table 4 shows an example of system combination.
The Chinese word {\em zuzhiwang} is an out-of-vocabulary(OOV) for NMT and the baseline NMT cannot correctly
translate this word. Although PBMT and HPMT translate this word well, they does not conform to the grammar.
By combining the merits of NMT and SMT, our model gets the correct translation.

\begin{table}
\begin{tabular}{p{1.22cm}p{0.8cm}<{\centering}p{0.8cm}<{\centering}p{0.8cm}<{\centering}p{0.8cm}<{\centering}p{0.8cm}<{\centering}}

  \hline
  System   & MT03      &   MT04      &  MT05      &   MT06   &  Ave \\
  \hline
  \hline
  NMT     & 1086      & 1145      &  1020     &  708  &  989.8  \\
  Ours     & \bf{869} & \bf{1023}  &  \bf{909} &  {\bf 609} & {\bf 852.5}  \\

  \hline
\end{tabular}
\caption{The number of unknown words in the results of NMT and our model.}
\end{table}

\begin{table}
\centering
\begin{tabular}{p{1.22cm}p{0.8cm}<{\centering}p{0.8cm}<{\centering}p{0.8cm}<{\centering}p{0.8cm}<{\centering}p{0.8cm}<{\centering}}
  \hline
  System   &    MT03   &   MT04   &  MT05   &   MT06   & Ave \\
  \hline
  \hline
  E-NMT     &   39.14    &   40.78  &  37.31   &   37.89   &   38.78   \\
  Jane    &   40.61     &   43.28   &  39.05   &   39.18   &   40.53   \\
  Ours    &   {\bf 43.61}   &   {\bf 47.65}   &  {\bf 42.02}   &   {\bf 41.17}   &   {\bf 43.61}    \\
  \hline
\end{tabular}
\caption{Translation results (BLEU score) when we replace original NMT with strong E-NMT, which uses ensemble strategy with four NMT models. All results of
  system combination are based on strong outputs of E-NMT.
  }
\end{table}

\subsection{Effect of Ensemble Decoding}

\begin{table*}
\centering
\begin{tabular}{p{1.5cm}p{9.5cm}}
  \hline
  Source & 海珊\ 也\ 与\ 恐怖\  \textcolor{red}{组织网} \ 建立\ 了\ 联系\ 。\\
  Pinyin &\textit{hanshan ye yu kongbu \textcolor{red}{zuzhiwang} jianli le lianxi 。} \\
  Reference & hussein has also established ties with terrorist networks .  \\
  \hline
  \hline
  PBMT & hussein also has established relations and terrorist group .   \\
  HPMT & hussein also and terrorist group established relations .   \\
  NMT &  hussein also established relations with UNK .  \\
  \hline
  Jane & hussein also has established relations with .  \\
  Multi & hussein also has established relations with the terrorist group . \\
  \hline
\end{tabular}
  \caption{Translation examples of single system, Jane and our proposed model.}
\end{table*}

The performance of candidate systems is very important to the result of system combination, and we use ensemble strategy with four NMT models
to improve the performance of original NMT system. As shown in Table 3, the E-NMT with ensemble strategy outperforms the original NMT
system by +1.40 BLEU points, and it has become the best sytem in all MT systems, which is +0.68 BLEU points higher than HPMT.

After replacing original NMT with strong E-NMT , Jane outperforms original result by +0.45 BLEU points,
and our model gets an improvement of +3.08 BLEU points over Jane.
Experiments further demonstrate that our proposed model is effective and robust for system combination.

\section{Related Work}

The recently proposed neural machine translation has drawn more and more attention. Most of the existing approaches and models mainly focus on
designing better attention models~\cite{Luong:2015A,Mi:2016A,Mi:2016B,Tu:2016,Meng:2016},
better strategies for handling rare and unknown words~\cite{Luong:2015B,Li:2016,Zhang:2016A,Sennrich:2016A} ,
exploiting large-scale monolingual data~\cite{Cheng:2016,Sennrich:2016B,Zhang:2016B},
and integrating SMT techniques~\cite{Shen:2016,Junczys-Dowmunt:2016B,He:2016}.

Our focus in this work is aiming to take advantage of NMT and SMT by system combination,
which attempts to find consensus translations among different machine translation systems.
In past several years,  word-level, phrase-level and sentence-level system combination methods were well studied
~\cite{Bangalore:2001,Rosti:2008,Li:2008,Li:2009,Heafield:2010,Freitag:2014,Ma:2015,Zhu:2016},
and reported state-of-the-art performances in benchmarks for SMT.
Here, we propose a neural system combination model which combines the advantages of NMT and SMT efficiently.

Recently, Niehues et al.~\shortcite{Niehues:2016} use phrase-based SMT to pre-translate the inputs into target translations.
Then a NMT system generates the final hypothesis using the pre-translation.
Moreover, multi-source MT has been proved to be very effective to combine multiple source languages
~\cite{Och:2001,Zoph:2016,Firat:2016a,Firat:2016b,Garmash:2016}.
Unlike previous works, we adapt multi-source NMT for system combination and design a good strategy
to simulate the real training data for our neural system combination.

\section{Conclusion and Future Work}

In this paper, we propose a novel neural system combination framework for machine translation. The central idea is to take advantage of NMT and SMT by adapting the multi-source NMT model.
The neural system combination method cannot only address the fluency of NMT and the adequacy of SMT, but also can accommodate the source sentences as input. Furthermore, our approach can further use ensemble decoding to boost the performance compared to traditional system combination methods.

Experiments on Chinese-English datasets show that our approaches obtain significant improvements over the best individual system and the state-of-the-art traditional system combination methods.
In the future work, we plan to encode n-best translation results to further improve the system combination quality. Additionally, it is interesting to extend this approach to other tasks like sentence compression and text abstraction.

\section*{Acknowledgments}

The research work has been funded by the Natural Science Foundation of China under Grant No. 61333018 and No. 61673380, and it is also supported by the Strategic Priority Research Program of the CAS under Grant No. XDB02070007.

\bibliography{acl2017}
\bibliographystyle{acl_natbib}

\appendix

\end{CJK*}
\end{document}